\documentclass[10pt,journal,compsoc]{IEEEtran}

\ifCLASSOPTIONcompsoc
  \usepackage[nocompress]{cite}
\else
  \usepackage{cite}
\fi

\usepackage{graphicx}
\usepackage{xcolor}
\usepackage{amsopn}
\usepackage{caption}

\begin{document}

\title{CaricatureShop: Personalized and Photorealistic Caricature Sketching}

\author{Xiaoguang~Han, Kangcheng~Hou, Dong~Du, Yuda~Qiu, Yizhou~Yu, Kun~Zhou, Shuguang~Cui 
	\IEEEcompsocitemizethanks{\IEEEcompsocthanksitem X. Han, Y. Qiu, and S. Cui are with Shenzhen Research Institute of Big Data, The Chinese University of Hong Kong, Shenzhen. K. Hou and K. Zhou are with Zhejiang University. D. Du is with University of Science and Technology of China and Shenzhen Research Institute of Big Data. Y. Yu is with The University of Hong Kong. \protect\\}
	\thanks{Manuscript received XXXX, 20XX}
}

%
%

\markboth{IEEE Transactions on Visualization and Computer Graphics,~Vol.~XX, No.~X, XXXX~20XX}%
{Han \MakeLowercase{\textit{et al.}}: CaricatureShop: Personalized and Photorealistic Caricature Sketching}
%


\IEEEtitleabstractindextext{%
\begin{abstract}
In this paper, we propose the first sketching system for interactively personalized and photorealistic face caricaturing. Input an image of a human face, the users can create caricature photos by manipulating its facial feature curves. Our system firstly performs exaggeration on the recovered 3D face model according to the edited sketches, which is conducted by assigning the laplacian of each vertex a scaling factor. To construct the mapping between 2D sketches and a vertex-wise scaling field, a novel deep learning architecture is developed. With the obtained 3D caricature model, two images are generated, one obtained by applying 2D warping guided by the underlying 3D mesh deformation and the other obtained by re-rendering the deformed 3D textured model. These two images are then seamlessly integrated to produce our final output. Due to the severely stretching of meshes, the rendered texture is of blurry appearances. A deep learning approach is exploited to infer the missing details for enhancing these blurry regions. Moreover, a relighting operation is invented to further improve the photorealism of the result. Both quantitative and qualitative experiment results validated the efficiency of our sketching system and the superiority of our proposed techniques against existing methods.
\end{abstract}

}

\maketitle

\IEEEdisplaynontitleabstractindextext

%


\ifCLASSOPTIONcompsoc
\IEEEraisesectionheading{\section{Introduction}\label{sec:intro}}
\else
\section{Introduction}
\label{sec:intro}
\fi


\begin{figure*}[t!]
	\includegraphics[width=\linewidth]{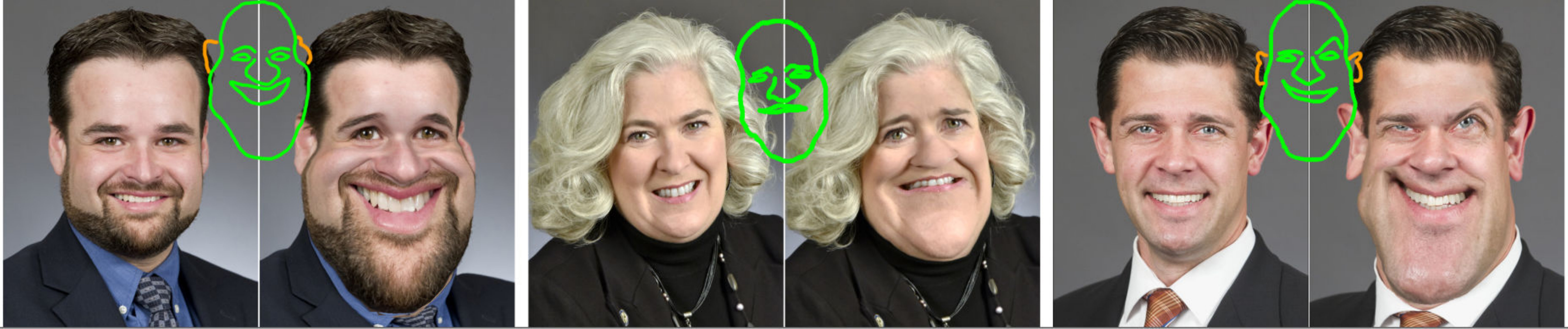}
	\centering
	\vskip -1mm
	\caption{Given a portrait photo, we propose a sketching system that allows users to manipulate facial feature curves for personalized and photorealistic caricaturing.}
	\label{fig:teaser}
	\vskip -3mm
\end{figure*}

\IEEEPARstart{D}{ue}
to the recent trend of integrating augmented reality with communication and social networking using smart devices, intelligent manipulation of human facial images, such as stylization\cite{fivser2017example}, expression editing\cite{averbuch2017bringing} and face reenactment\cite{thies2016face2face}, has become widely popular. However, there has been relatively little work on applying large geometric deformations to faces that appear in images, including making caricatures from real face pictures. Yet, most of such work~\cite{liang2002example, liao2004automatic} convert face pictures into caricatures in the line drawing style.
In this paper, we focus on sketch-based creation of personalized caricatures that have photorealistic shading and textures. Such high-quality photorealistic caricatures represent an important form of augmented reality.

There are three requirements in our goal, sketch-based interaction, personalization and photorealism. Since there exists a large degree of freedom when caricatures are created, interactive guidance is necessary. By mimicking free-hand drawing, sketching is a form of interaction that is both efficient and user-friendly. Personalization is important since users prefer to preserve the identities of faces when making caricatures. Photorealism is made possible by following the commonly adopted 3D-aware strategy that exploits an underlying 3D face model recovered from the input face image.


Satisfying the above three requirements while making caricatures is challenging for the following reasons. First, sketching only provides very sparse inputs while creating an expressive caricature requires dense edits over the facial surface. The mapping between them is highly non-linear. It is nontrivial to design an effective machine learning algorithm to learn such a mapping, which not only needs to be very sensitive to local shape variations but also guarantees the dense edits still keep the original face identity recognizable. Second, during the process of creating a caricature, facial geometry experiences large deformations, incurring local geometry changes, which gives rise to inconsistencies between the altered geometry and the original appearance properties, including incorrect shading and insufficient texture resolution in stretched regions. Third, while we can recover a 3D face model for the facial region in the input image, we do not have 3D information for the rest of the facial image, including hair and other parts of the human body. During image re-rendering, how can we warp the image regions without 3D information so that they become consistent with the facial region with 3D information?

In this paper, we propose a novel sketch-based system for creating personalized and photorealistic caricatures from photographs.
Given an input image of a human face and its underlying 3D model, reconstructed using the approach in \cite{cao2014facewarehouse}, our system can produce a photorealistic 2D caricature in three steps: texture mapping the 3D model using the input image, 3D textured face exaggeration, re-rendering the exaggerated face as a 2D image. To tackle the core problem of sketch-based face exaggeration, we introduce a deep learning based solution, where the training data consists of encodings of the input normal face models, the input sketches and their corresponding exaggerated 3D face models. Since meshes have irregular connectivities unsuitable for efficient deep learning algorithms, these encodings are defined as images over a 2D parametric domain of the face, and face exaggeration is cast as an image-to-image translation problem~\cite{isola2017image}. To support network training, a large synthetic dataset of sketch-to-exaggeration pairs is created.

We also propose effective solutions to address technical problems encountered during face image re-rendering. First, to fix incorrect shading effects due to facial geometry changes, an optimization algorithm is developed to find an optimal pixel-wise shading scaling field. Second, insufficient texture resolution caused by face exaggeration usually makes certain local regions in the re-rendered image blurry. Deep learning based image-to-image translation is exploited again to handle this problem by learning to infer missing high-frequency details in such blurry regions. To achieve efficient performance required by our sketching interface, we divide the input photo into overlapping patches and run a lightweight pix2pix network~\cite{isola2017image} on individual patches separately. To avoid seams along patch boundaries, the deep network is trained for inferring high-frequency residuals instead of final pixel colors. Third, inconsistencies between regions with underlying 3D model and those without result in artifacts especially on facial boundaries, ears and hair regions. To remove such artifacts, we first generate two images, one obtained by applying 2D warping guided by the underlying 3D mesh deformation and the other obtained by re-rendering the deformed 3D textured model. These two images are then seamlessly integrated to produce our final output.


\textbf{Contributions.} In summary, this paper has the following contributions:
\begin{itemize}
	\item We propose a comprehensive easy-to-use sketching system for interactive creation of personalized and photorealistic caricatures from photographs. Our system is made possible by a suite of novel techniques for 3D face exaggeration, exaggerated face re-shading, image detail enhancement, and artifact-free caricature synthesis.
	\item We design a novel deep learning based method for inferring a vertex-wise exaggeration map for the underlying 3D face model according to user-supplied 2D sketch edits.
	\item A deep neural network for patch-oriented residual inference is devised to infer additional high-frequency details for improving the resolution of stretched textures during re-rendering.
	\item Two datasets are built for training and testing deep neural networks used in our sketching system. The first one is a large synthetic dataset for training the deep network that maps sparse sketches to a dense exaggeration map. The second one is a dataset of high-resolution (~$1080p$ and above) portrait photos for training the deep network that synthesizes high-frequency details for facial textures with insufficient resolution. These datasets will be publicly released to benefit other researchers working in this area.
\end{itemize}



\section{Related Work}
\label{sec:related_work}

We study the literature reviews from the following four aspects.

\begin{figure*}[t]
	\centering
	\includegraphics[width=0.95\textwidth]{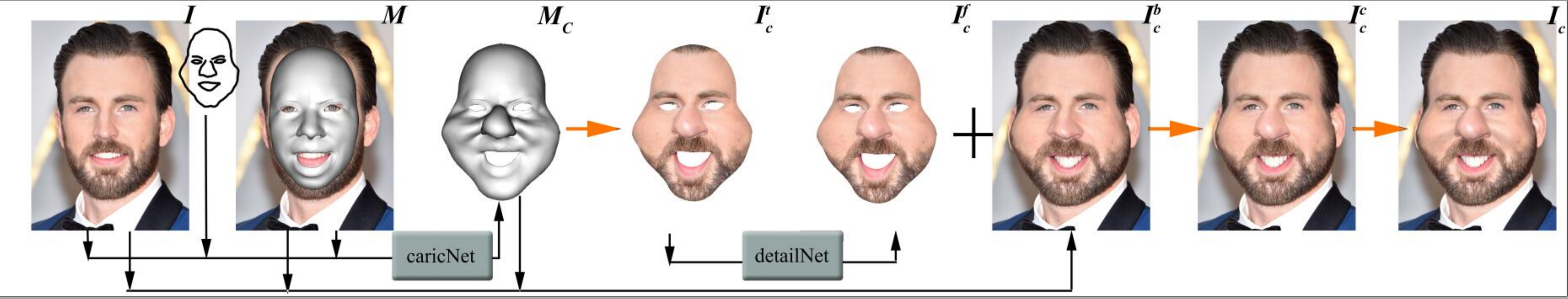}
	\caption{The framework of our method. Taking an image of a human face and an edited sketch as input, it tries to output a photorealistic photo in accordance with the sketch. }
	\label{fig:overview}
\end{figure*}

\textbf{2D Caricature Generation.}
Input an image of a human face, to create its caricatures using computer algorithms can be dated back to the work of \cite{brennan1985caricature} which presented the first interactive caricature generation system. Akleman \cite{akleman1997making} further proposed an interactive tool with morphing techniques. Afterwards, many approaches tried to automate this procedure. For example, Liang et al. \cite{liang2002example} developed an automatic approach to learn exaggeration prototypes from a training dataset. Based on the prototype, shape exaggeration and texture style transfer were then applied to create final results. The work of ~\cite{liu2003efficient} proposed to learn an Inhomogeneous Gibbs Model (IGM) from a database of face images and their corresponding caricatures. Using the learnt IGM, caricatures can be generated automatically from input photos. By analyzing facial features, Liao et al. \cite{liao2004automatic} invented an automatic caricature generation system using caricature images as references. We highly recommend the readers referring to \cite{sadimon2010computer} for a detailed survey of computer-aided caricature generation. Different from these works who are aiming to create caricatures in abstract line styles, in this paper, we focus on the generation of phtorealisic caricatures.

\textbf{3D Caricature Modeling.}
There also exists many works on 3D caricature model creation from a normal 3D face model. This is commonly conducted by firstly identifying the distinctive facial features and then exaggerating them using mesh deformation techniques. Both \cite{lewiner2011interactive} and \cite{vieira2013three} performed exaggeration on an input 3D model by magnifying its differences with a template model. The gradients of vertices on the input mesh are considered as a measure of facial peculiarities in \cite{sela2015computational}. The exaggeration is thus carried out by assigning each vertex a scaling factor on its gradient. In addition to, some works attempt to modeling 3D caricatures from images. For example, Liu et al. \cite{liu2009caricature} developed a semi-supervised learning approach to map facial landmarks to the coefficients of a PCA model learned from a set of 3D caricatures. Wu et al. \cite{wu2018alive} introduced an intrinsic deformation representation that enables large face exaggeration, with which an optimization framework was proposed to do 3D reconstruction from a caricature image. Recently, with advanced deep learning techniques, Han et al. \cite{han2017deepsketch2face} proposed deepsketch2face which trained a Convolutional Neural Networks (CNNs) to build the mapping from 2D sketches to the coefficients of a bilinear morphable model representing 3D caricatures. Although we also targets to producing 3D caricature models from 2D sketches, it differs from this method in two aspects. Firstly, the output of deepsketch2face only related to the sketches while our result depends on both the input 3D face model and the manipulated sketches. This makes our approach targeting personalized caricaturing, i.e., different faces with same sketches can have different results. Secondly, deepsketch2face utilizes a 66-dimensional vector to represent 3D caricature shape space while our method uses vertex-wise scaling factors for the representation which gives rise to larger space.

\textbf{3D-aware Face Retouching.}
With the rapid development of 3D face reconstruction from a single image (e.g., \cite{blanz1999morphable, kemelmacher20113d, cao20133d, jackson2017large}), a large body of works have validated that 3D facial information can greatly help intelligent face retouching. For example, Yang et al. \cite{yang2011expression} proposed an approach to transfer expressions between two portrait photos with same facial identities. To do this, the 3D models of the two input images are firstly recovered and their deformations are then projected to produce a warping field. This method is then utilized in \cite{yang2012facial} for expression editing of facial videos. Such 3D-aware warping strategy is also adopted by \cite{fried2016perspective} to simulate the changing of the relative pose and distance between camera and the face subject. Shu et al. \cite{shu2017eyeopener} also took advantage of this warping method to make the closed eyes in photographs be open. Using a re-rendering framework, the works of \cite{thies2015real} and \cite{thies2016face2face} successfully developed systems for real-time facial reenactment on videos. To the best of our knowledge, we are the first work to perform very large geometric deformations on images. This will cause: a) self-occlusions; b) visually-unreasonable shading effects; c) blurry texturing. These make existing methods fail. In this paper, several techniques are designed to deal with such problems as described in Section \ref{sec:texture} and Section \ref{sec:synthesis}.

\textbf{Facial Detail Enhancement.}
The details are usually missing when the portrait images undergo downsampling. To infer the missing details and produce high-resolution photos from low-resolution ones, also called face hallucination, is one of the most popular topics in computer vision recently. The readers can refer to \cite{wang2014comprehensive} for a detailed survey of this area. Here, we only give a literature review of face hallucination methods based on deep learning architectures. A discriminative generative network is firstly introduced in \cite{yu2016ultra} for super-resolving aligned low-resolution face images. In \cite{yu2017face} and \cite{yu2017hallucinating}, the same authors proposed to involve spatial transformer networks to the deconvolutional layers for dealing with unaligned input images. FSRNet, developed in \cite{chen2017fsrnet}, leveraged facial parsing maps as priors to train the deep network in an end-to-end manner. Bulat and Tzimiropoulos \cite{bulat2017super} further improved the performance by performing face super-resolution and facial landmark detection simultaneously in a multi-task learning way. All of these methods only take low-resolution images as input while our sketching system allows high resolution input. Although the method of \cite{wang2017high} is able to take over our settings, the proposed neural networks are too heavy to support efficient user interactions. Our work tackles this issue by using a patch-based learning approach which is described in Section \ref{sec:texture}.

\section{Overview}
\label{sec:overview}

Our system takes a single image of a human face as input which is denoted as $I$. The method in \cite{cao20133d} is first applied to obtain a 3D face mesh $M$ that captures both the identity and expression. As in \cite{han2017deepsketch2face}, a set of pre-defined feature curves on $M$ are rendered as 2D sketched lines for manipulation. The details of our sketch editing mode will be introduced in Section~\ref{subsec:ui3dex}. The manipulated sketch lines together with $M$ are fed into our deep learning based 3D caricaturing module to create a caricature model $M_{c}$. This process will be described in Sections~\ref{subsec:dl3dex} and~\ref{subsec:exp3dex}. The next step synthesizes a photorealistic caricature image $I_{c}$, and consists of three phases. First, $M_{c}$ is re-rendered using the texture map of $M$ to create image $I^{t}_{c}$ ($t$ is short for texture). Note that rendering $M_{c}$ with the original texture map of $M$ usually produces blurry regions in $I^{t}_{c}$ due to severe stretching caused by exaggeration in certain local regions of $M_{c}$. A deep neural network is used to enhance such regions of $I^{t}_{c}$ by inferring missing high-frequency facial details. We denote the enhanced $I^{t}_{c}$ as $I^{f}_{c}$ ($f$ means foreground). The details of this phase will be elaborated in Section~\ref{sec:texture}. Second, $I$ is warped according to the 3D deformations that transform $M$ to $M_{c}$, and the result is denoted as $I^{b}_{c}$ ($b$ means background). Third, $I^{f}_{c}$ and $I^{b}_{c}$ are fused together to output $I_{c}$. Image fusion consists of two substeps: they are first stitched together by finding an optimal cut, and then a relighting operation eliminates inconsistent shading effects. Technical details of the last two phases will be discussed in Section~\ref{sec:synthesis}. In Section~\ref{sec:synthesis}, we will also describe an interactive refinement module for mouth region filling and sketch-based ear editing. The complete pipeline of our system is illustrated in Fig~\ref{fig:overview}.

\section{Sketch-Based Caricaturing}
\label{sec:exaggeration}

\begin{figure}
  \centering
  \includegraphics[width=0.45\textwidth]{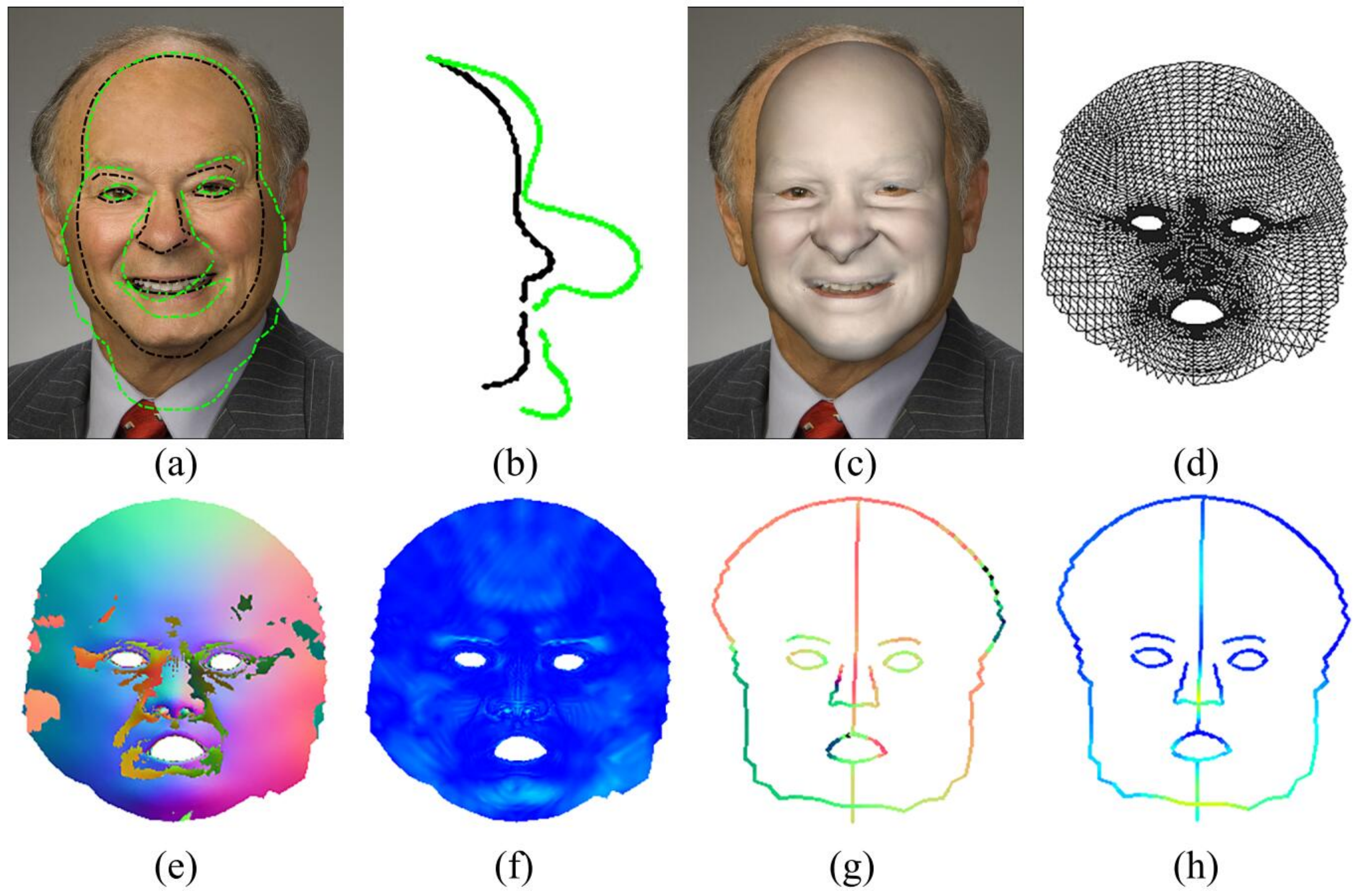}
  \caption{We flatten both the shape and sketch information into a parametric domain. (a) shows an input image and the sketches before (in red) and after (in green) users' editing. (b) shows the editing in profile view. (c) shows the recovered 3D shape. (d) gives the parametric domain used for our information flattening. (e)-(h) show $L_d, L_m, S_d, S_m$ respectively.}
  \label{fig:infoflattening}
\end{figure}

In this section, we describe the details of our sketch-based caricaturing method, that performs shape exaggeration on $M$ to obtain $M_{c}$. Note that when we apply the method in \cite{cao20133d} to recover the 3D face model $M$ from the input image, another face model $M^0$ that has a neutral expression but shares the same identity as $M$ can also be obtained. Our caricaturing process has two phases: a) identity exaggeration, which obtains an exaggerated neutral face model $M^{0}_{c}$ by applying distortions to deform $M^{0}$; b) restore the expression of $M$ on the exaggerated neural face $M^{0}_{c}$ to obtain $M^{e}_{c}$. Then we follow the practice in \cite{han2017deepsketch2face} to obtain the final output of our method $M_{c}$ by further performing handle-based mesh deformation on $M^{e}_{c}$ to make the model exactly match the drawn sketch lines.

\subsection{User Interface}
\label{subsec:ui3dex}
Let us first briefly introduce our user interface. Our basic sketching interface is similar to the follow-up sketching mode in \cite{han2017deepsketch2face}. Specifically, the silhouette line of the face region and a set of feature lines (i.e., contours of mouth, nose, eyes and eyebrows) on $M$ are projected and overlayed on the input photo, as shown in Fig~\ref{fig:infoflattening} (a). These 2D sketch lines are displayed for manual manipulation. An erase-and-redraw mechanism is used for sketch editing: once an erasing operation has been performed on any silhouette or feature line, a drawing operation is required to replace the erased segment with a new one. The silhouette line is represented as a closed curve consisting of multiple connected segments. When one of its segments has been redrawn, auto-snapping is applied to remove the gap between end-points. To ensure a friendly editing mode, all user interactions are performed from the viewpoint of the input photo, which can also be recovered using the method in \cite{cao20133d}. Our user interface differs from that in \cite{han2017deepsketch2face} in two aspects. First, we provide an additional option for users to edit sketches in a side view. This makes the modification of some feature lines much easier. Moreover, the feature lines around ears can be manipulated in our system because misalignments between $I$ and $M$ in the regions around ears give rise to artifacts negatively affecting further image synthesis. We leave sketch-based ear editing as a refinement module, which will be discussed in Section~\ref{sec:synthesis}.

From the perspective of a user, he/she first loads a face image. The editing mode is then started by a button click and the 2D sketch lines are displayed immediately. Thereafter, the user can manipulate the sketch lines according to a mental image of the caricature the user plans to make. During this process, the user can switch to the side view at any time. Such switching triggers our 3D face exaggeration engine and the sketch will be updated according to the latest exaggerated model. The same happens when the frontal view is switched back.

\subsection{Identity Caricaturing}
\label{subsec:dl3dex}
To be convenient, we assume the input photo is a frontal view of a face and the user performs editing in this view. Discussions on other cases are left to the end of this section. We denote the sketches before and after editing as $S$ and $S_c$, respectively. In the following, we describe how to generate $M^{0}_c$ from both $M^{0}$ and $S_c$.

\textbf{Mesh Exaggeration.}
We perform mesh exaggeration by following the idea in \cite{sela2015computational}, which assigns each mesh vertex an exaggeration factor. Given the original mesh $M^{0}$ represented using a set of vertices $V$ and a set of edges $E$, for each vertex $v_i$, we scale its Laplacian $\delta_i$ with an exaggeration factor $\lambda_i$. The readers are referred to \cite{nealen2006laplacian} for the definition of $\delta_i$. The coordinates of all vertices over the exaggerated mesh $M^{0}_c = (V', E')$ are calculated by solving a sparse linear system. This system is built using Laplacian constraints $\delta'_i = \lambda_i\delta_i$ at all vertices and position constraints at a small number of vertices on the backfacing part of the face model. Thus, the problem of creating the exaggerated mesh $M^{0}_c$ is transformed to the problem of defining exaggeration factors for all vertices.

\begin{figure}
  \centering
  \includegraphics[width=0.45\textwidth]{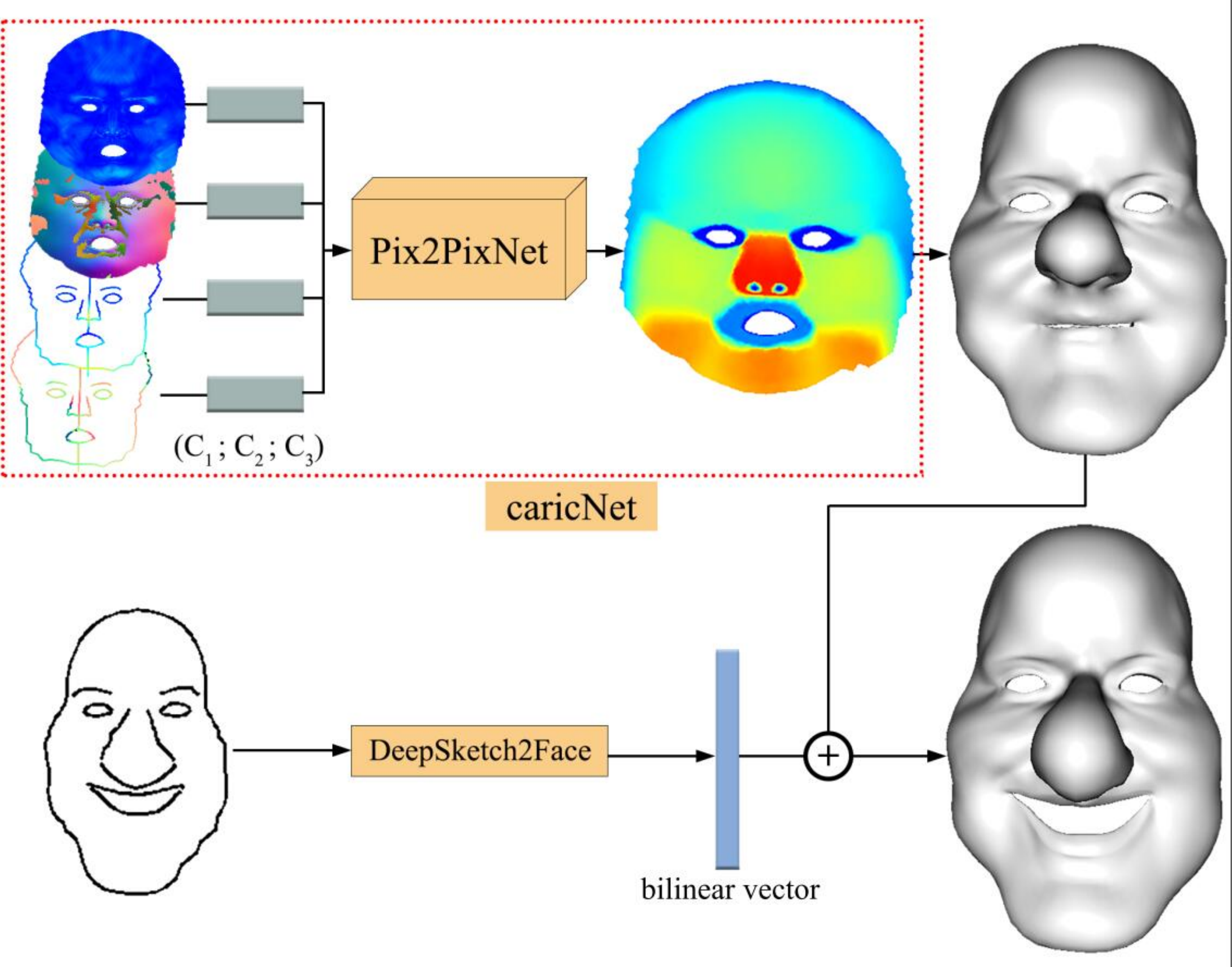}
  \caption{The pipeline of our 3D caricaturing. We used caricNet for exaggeration generation and used DeepSketch2Face for expression inference, which are then combined to produce final result. The input maps are followed by four individual streams of CNNs. All the streams own same architecture. It is consisting of 3 Conv+Relu layers, denoted as $(C1;C2;C3)$. $C1$ is of 5*5 Conv kernel and 128 channels. $C2$ is of 3*3 Conv kernel and 64 channels. $C3$ is of 3*3 Conv kernel and 32 channels. }
  \label{fig:caricPipeline}
\end{figure}

\textbf{Information Flattening.}
It is obvious that how much a vertex needs to be exaggerated highly depends on both the shape of the edited sketch lines and the characteristics of the face geometry. Therefore, exaggeration factors can be defined by mapping $M^{0}$ and $S_c$ to vertex-wise $\lambda$'s. Although deep learning algorithms are well suited for approximating such non-linear mappings, there are no effective deep network architectures for vertex-wise regression over meshes because of their irregular connectivity. Nonetheless, due to the consistent topology of face models, it is feasible to use a 2D parameterization of an average neutral face (denoted as $M^{0}_a$) as a common domain, where flat 2D maps (images) can represent dense vertex-wise information over face meshes. Exaggeration factors for all vertices can thus be represented using an image by first assigning every projected vertex in the parametric domain a color that encodes its exaggeration factor and then interpolating the colors at all other pixels. This image is called $\lambda$ map. The input mesh $M$ can be similarly embedded into this parametric domain. Instead of encoding the position of vertices, in our method, the Laplacian is used to represent $M$. This is because the Laplacian has a close connection with the $\lambda$ map and a mesh can be reconstructed from its vertex-wise Laplacians and an extra boundary condition. As the Laplacian of a vertex is a vector, we encode its direction and magnitude separately in two maps, a direction map ($L_d$) and a magnitude map ($L_m$). The process to represent $S_c$ has three steps: 1) we first embed 3D feature lines on the exaggerated face mesh into the parametric domain to define a 2D sketch $S^{p}_c$; 2) we project the 3D feature lines on $M^{0}_a$ into the image plane of $I$ and produce another sketch $S^{0}_a$; 3) for every point on $S^{p}_c$, we calculate the displacement vector between its corresponding points on $S^{0}_a$ and $S_c$. The vectors defined on $S^{p}_c$ are also decomposed into a direction map $S_d$ and a magnitude map $S_m$. Thus, the problem becomes computing a mapping between $(S_d, S_m, L_d, L_m)$ and the $\lambda$ map (all maps are shown in Fig~\ref{fig:infoflattening}).

\begin{figure}
  \centering
  \includegraphics[width=0.45\textwidth]{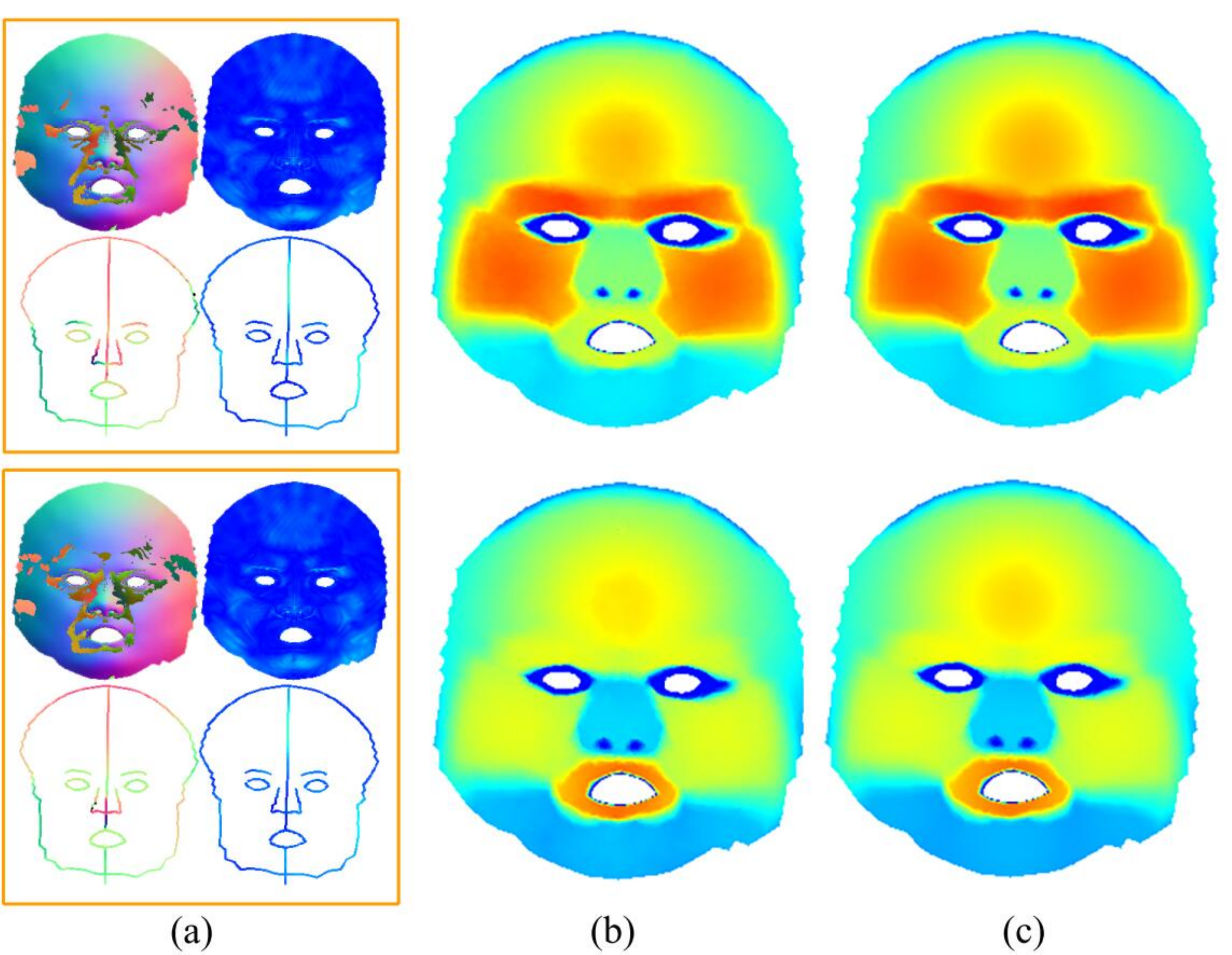}
  \caption{Two results of our caricNet for $\lambda$ map inference. (a) shows the input. (b) shows the output $\lambda$ maps of caricNet and (c) shows its groundtruth. }
  \label{fig:lamda}
\end{figure}

\textbf{Pix2Pix-Net.}
Before introducing our network design, let us have a review of the Pix2Pix-Net \cite{isola2017image}, which is designed to learn the transformation between two images in different styles yet having similar spatially distributed contents. It consists of two sub-networks: a generative network (G-net) and a discriminator network (D-net). The basic architecture of the G-net is U-net \cite{ronneberger2015u}. It takes an input $x$ and produces an output $y$. Inside this network, the input passes through a series of convolutional and pooling layers until a bottleneck layer, from which point the process is reversed. There are skip connections between every pair of corresponding layers in the first and second halves of the network. At the second endpoints of these connections, all channels are simply concatenated. This design makes full use of the low-level information in the input by fusing them with layers close to the output. To improve the visual quality of the output, the D-net tries to learn a classifier to identify all results produced by the G-net. Generative adversarial networks (GANs) are generative models that learn a mapping from a random noise vector $z$ to an output image $y$~\cite{goodfellow2014generative} while conditional GANs learn a mapping from an input $x$ and a random noise vector $z$ to an output $y$. The loss function for training Pix2Pix-Net are defined below:
\begin{equation}\label{eq:lall}
  L_{all} = \arg \min_{G} \max_{D} L_{cgan}(G,D)+\lambda L_1(G),
\end{equation}
where $L_1$ measures the distance between the output and the ground truth, and
\begin{equation}\label{eq:lcgan}
  L_{cgan} = E_y [ \log D(y)] + E_{x,z} [\log (1-D(G(x, z)))].
\end{equation}

\begin{figure}
  \centering
  \includegraphics[width=0.45\textwidth]{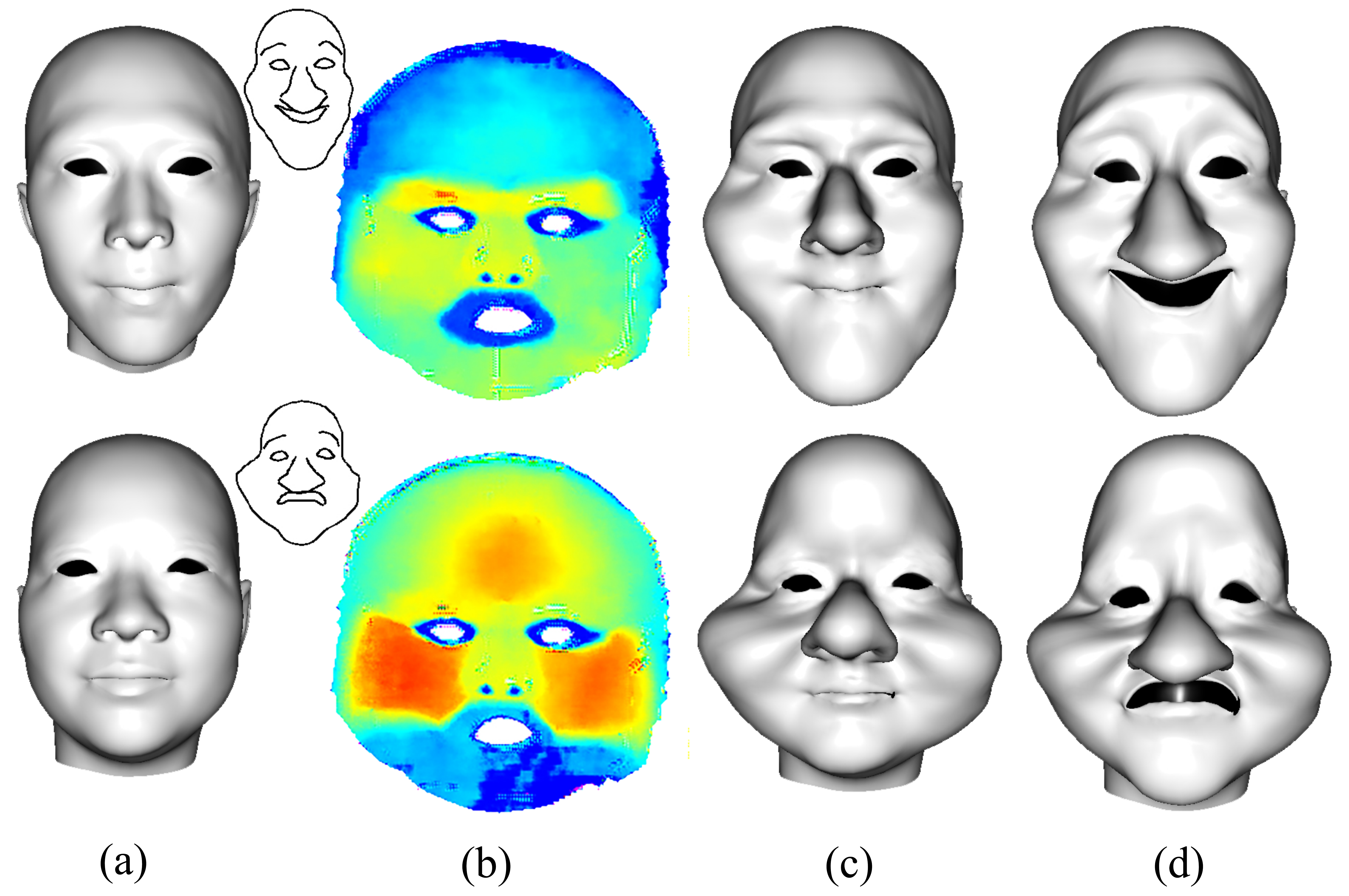}
  \caption{Two results of our 3D caricaturing. (a) shows the input neutral models with edited sketches over them. (b) shows the $\lambda$ map inferred using our caricNet. (c) shows the result of identity caricaturing. (d) shows final result with expression modeling and the handle-based deformation.}
  \label{fig:exgresults}
\end{figure}

\textbf{Our Network.}
As shown in Fig~\ref{fig:caricPipeline}, our network takes four images (i.e., $S_d$, $S_m$, $L_d$, $L_m$ ) as input. These four images are fed into four branches of convolutional layers, respectively. Each branch has three convolutional layers with ReLu activation. The output feature maps from these branches are then concatenated and fed into a Pix2Pix-Net for producing the $\lambda$ map. We call this network {\em caricNet}. Here, we do not directly concatenate the four input images together to form one image with multiple channels since this results in worse results in our experiments. One possible reason is that the four images are defined in very different information domains and our design tries to transform them into similar feature spaces before concatenation. Our training loss follows (\ref{eq:lall}) and (\ref{eq:lcgan}) by replacing $x$ with $(S_d, S_m, L_d, L_m)$. From the generated $\lambda$ map, we can simply take the value for each vertex on $M^{0}$ and perform exaggeration by solving a linear system to obtain $M^{0}_c$. Note that $S_c$ is a sketch that includes information about facial expression. Our network is trained to treat such information as noise, and only infers the exaggeration map for the face model with a neutral expression. Explicit expression modeling will be considered later in Section~\ref{subsec:exp3dex}.

\textbf{Dataset and Training.}
To support the training of our deep network, we build a large synthetic dataset of $(S_d, S_m, L_d, L_m)$ and their corresponding $\lambda$ maps. At first, we take the 150 neutral face models with distinct identities from Facewarehouse~\cite{cao2014facewarehouse} and apply 10 different types of exaggerations to each individual model. To carry out each exaggeration, we divide a face model into several semantic regions, including forehead, chin and nose. A random scaling factor is assigned to the center vertex of each region, and scaling factors for all other vertices in the same region are set with a Gaussian kernel. Once the exaggeration factors for all vertices have been assigned, they are smoothed to ensure the smoothness of the exaggerated model. 1,500 neutral caricature models are thus created. The 25 expressions used in \cite{han2017deepsketch2face} are transferred to each of these models using the algorithm in \cite{sumner2004deformation}. In total, we generate $37,500$ models. The 3D feature curves on these models are then projected to produce corresponding 2D sketches. The inferred $\lambda$ maps of our \emph{caricNet} associated with their goundtruth for two examples are shown in Fig \ref{fig:lamda}.

\textbf{Discussion on Viewpoints.}
We address two questions about viewpoints here. First, if the input face photo is not exactly a frontal view, we apply an affine transformation to make it frontal and then use our CaricatureNet for exaggeration inference. Second, given a sketch from the side view, we can also represent feature curves in the parametric domain, and colors assigned to points on these curves encode the displacements of the manipulated sketch lines. Such displacement vectors can also be decomposed into a direction map $S_d$ and a magnitude map $S_m$. Moreover, we train an additional network, called \emph{caricNet-side}, using a dataset that contains side-view sketches. During a sketching session, we use \emph{caricNet-side} when the side-view sketching mode is triggered, and \emph{caricNet} otherwise.

\subsection{Expression Modeling}
\label{subsec:exp3dex}
In this section, we explain how to generate $M^{e}_c$ from $M^{0}_c$ and $S_c$ by modeling facial expressions. Our method has two steps: expression regression from $S_c$ and expression transfer to obtain $M^e_c$. Specifically, we first use the $1,500$x$25$ models to train a bilinear morphable model as in \cite{cao2014facewarehouse}. This morphable model represents a 3D face model using a 50-dimensional identity vector (denoted as $u$) and a 16-dimensional expression vector (denoted as $v$). Given such a representation, we follow the same practice as in \cite{han2017deepsketch2face} to train a CNN based regression network that maps $S_c$ to both $u$ and $v$ while inferring a 3D model $M_r$ simultaneously. Here, we also train the network for $S_c$ using images with frontal views, and input sketches are transformed to frontal ones  before expression regression. To produce $M^{e}_c$, the expression of $M_r$ is transferred to $M^{0}_c$ using the deformation transfer algorithm from \cite{sumner2004deformation}. In Fig \ref{fig:exgresults}, two examples are used to show the procedure of our 3D caricaturing. 

\section{Facial Detail Enhancement}
\label{sec:texture}

\begin{figure}
  \centering
  \includegraphics[width=0.45\textwidth]{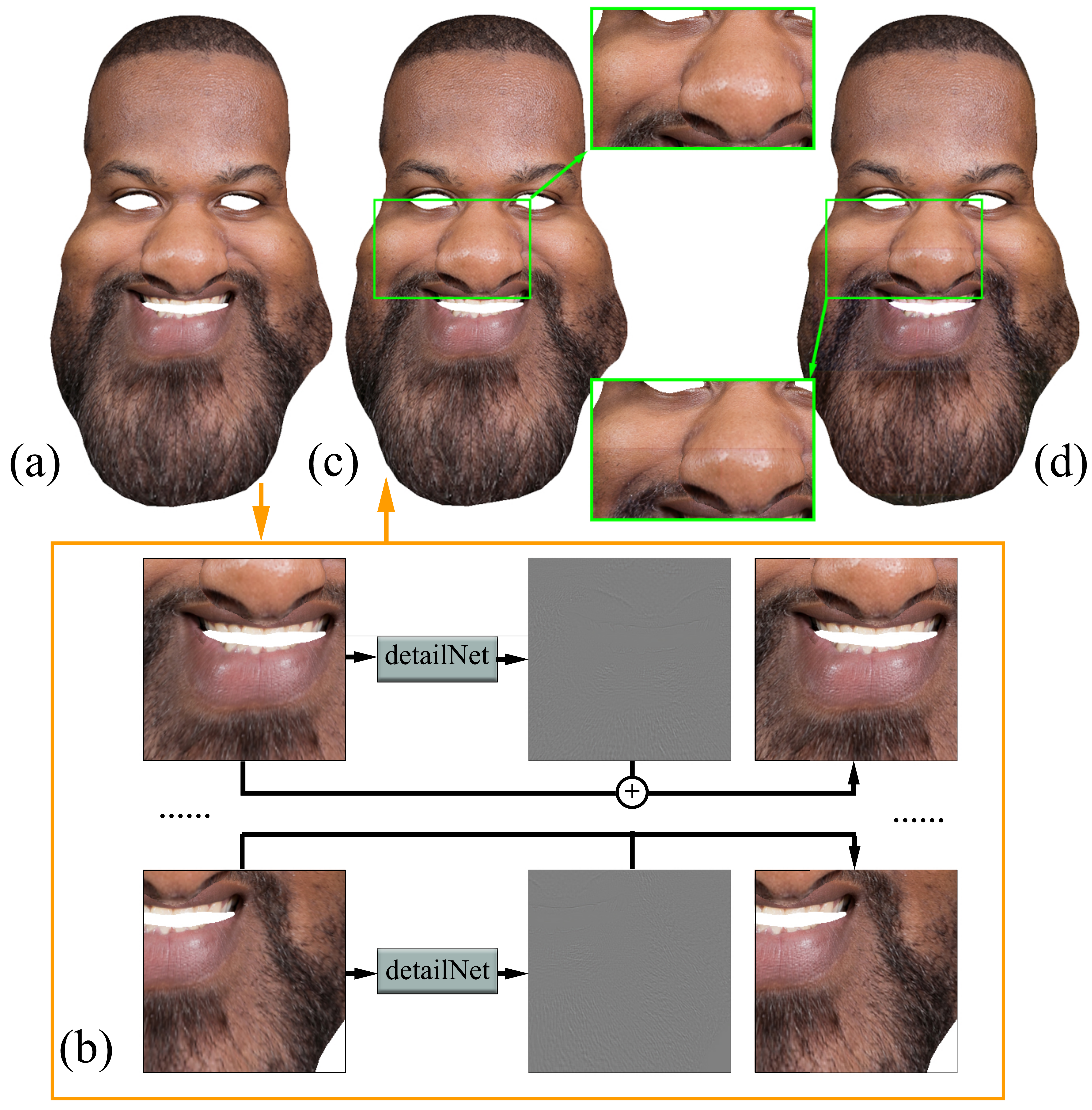}
  \caption{Taking an image (a) as input, our approach split it into several overlapped patches. Each patch is enhanced using our \emph{detailNet} as shown in (b). Both the residual and the final result for each patch are shown. Finally, we obtain the result (c) with detailed appearances. (d) shows the result without using residual which causes seams.}
  \label{fig:detailenhance}
\end{figure}

\begin{figure}
  \centering
  \includegraphics[width=0.45\textwidth]{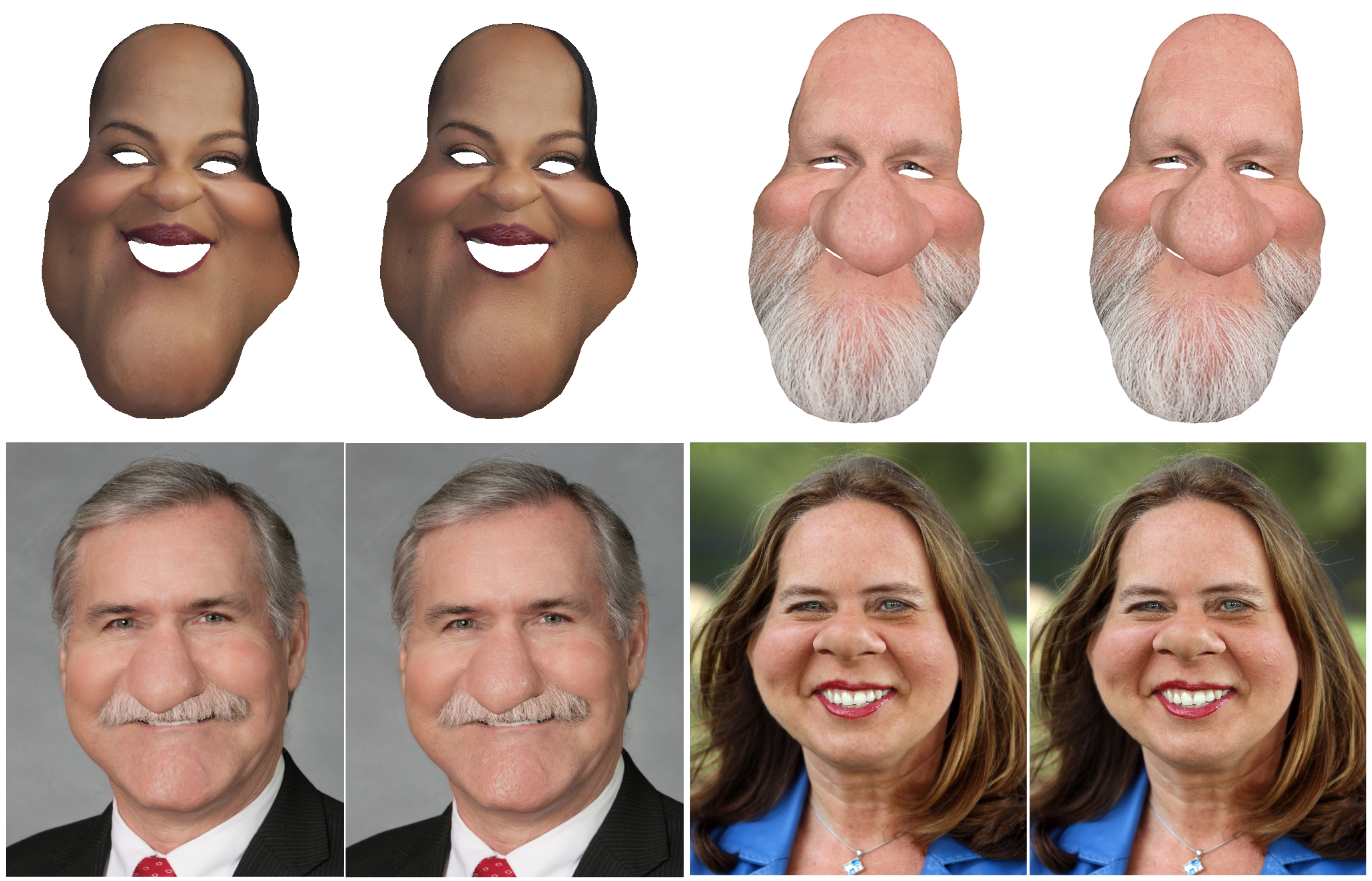}
  \caption{More examples of our detail enhancement. For each example, the results before and after enhancing are shown. The images in the second row are shown with their background. }
  \label{fig:detailresult}
\end{figure}

Given the correspondences between $M$ and $I$, we can directly perform texture mapping to generate a textured model for $M_c$. This model can be re-rendered to produce an image patch serving as the foreground content for our final result $I_c$. However, mesh exaggeration makes some triangle faces undergo severe stretching, which results in insufficient texture resolution and blurry appearances as illustrated in Fig~\ref{fig:detailresult}. We propose a deep learning based method for detail enhancement.

\textbf{Method.}
A method for enhancing texture resolution is necessary in order to show high frequency details such as beard and freckle. Deep learning techniques have been proven very good at such face hallucination problems~\cite{yu2017face,chen2017fsrnet, bulat2017super}. In our work, we use the Pix2Pix-Net again for translating blurry input images to their resolution enhanced versions. Unfortunately, this network becomes less efficient and the running speed is too slow when the input image has a large size. This performance issue limits the usability of our system as the users often wish to process high-resolution inputs. This problem is addressed in a divide-and-conquer manner. We first divide an input image into several overlapping patches and then perform detail enhancement over individual patches. A patch-level Pix2Pix-Net, which takes a patch with a fixed size as input, is trained. However, as shown in Fig~\ref{fig:detailenhance}, this strategy gives rise to seams due to the lack of consistency along patch boundaries. To tackle this problem, instead of straightforwardly inferring the detail-enhanced patch $P_s$ from the blurry input $P_b$, our network is trained to predict the residual $P_s-P_b$, which represents high-frequency details only. As such details themselves are not very spatially coherent, the seams between adjacent patches are naturally eliminated. This is demonstrated in Fig~\ref{fig:detailenhance}. In detail, our network takes a $256$x$256$ patch $p$ as input and produces a high-frequency residual $r$. The Pix2Pix-Net is also exploited to approximate the mapping from $p$ to $r$. We denote this network as \emph{detailNet}.

\textbf{Dataset.}
To the best of our knowledge, there are no datasets containing high-resolution face photos available for our problem. Therefore, we built our own dataset by collecting high-resolution facial portraits. We manually collect 899 photos ranging from $720p$ to $2K$ (after a crop to make the face region fill at least half of the image). Afterwards, for each image in the dataset, we first apply the method in \cite{cao20133d} to recover its 3D face model, which is then exaggerated into 10 different levels. Given an exaggerated model $X$ and its corresponding input image $I$, we create a pair of blurry photo $I_b$ and its detail-enhanced version $I_s$ in the following three steps. At first, we obtain a downsampled image $I_d$ from an original image $I$ according to the exaggeration level, which is measured by the average scaling factor of all faces in $X$. Second, $X$ is texture mapped using $I_d$, and then projected into an image region with the same size as $I_d$. This produces $I_b$. Third, to create $I_s$, $X$ is texture mapped using $I$, and also projected into an image region with the same size as $I_b$. In total, we generate 8,990 pairs of images like $I_b$ and $I_s$. 10 patches are randomly cropped from each pair to form a network training set.

\section{Caricature Photo Synthesis}
\label{sec:synthesis}
\begin{figure}
  \centering
  \includegraphics[width=0.45\textwidth]{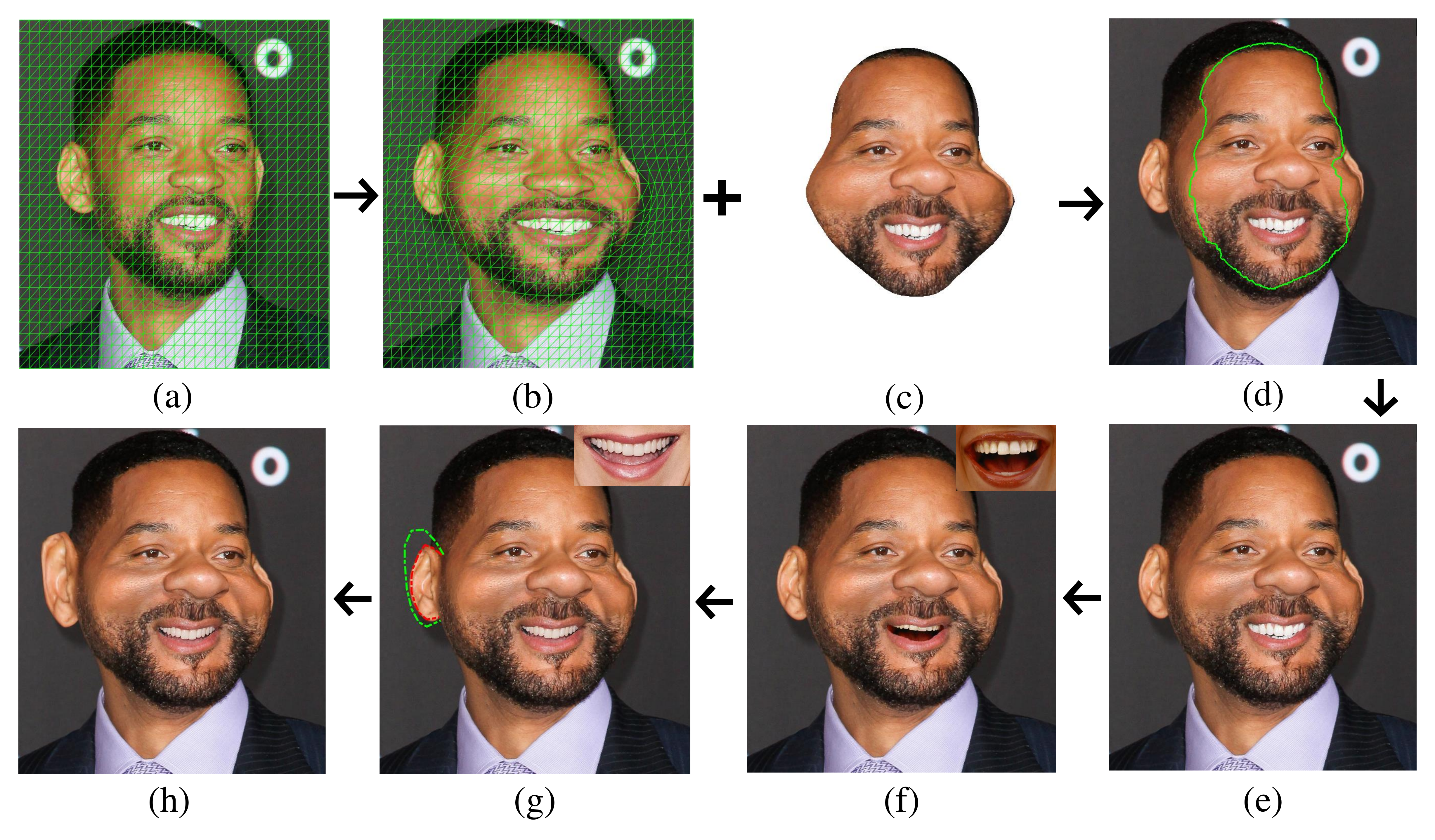}
  \caption{For photo synthesis. The input image (a) is firstly warped to (b) using the mesh-based image deformation strategy. With the enhanced rendered image (c), we fuse it with (b) together by finding an optimal cut (in green) as shown in (d). This is then followed by a reshading method and generate (e). Then, we can do interactively refinement by replacing mouth regions as shown in (f) and (g) and ear editing as shown in (h).}
  \label{fig:synthesis_pipeline}
\end{figure}

Given a portrait photo $I$, in previous sections, we have explained the procedure of generating a foreground image $I^{f}_c$. However, it only provides the content for pixels in the frontal regions of the naked 3D face model. We discuss how to synthesize the final caricature photo in the following two subsections.

\begin{figure}
  \centering
  \includegraphics[width=0.45\textwidth]{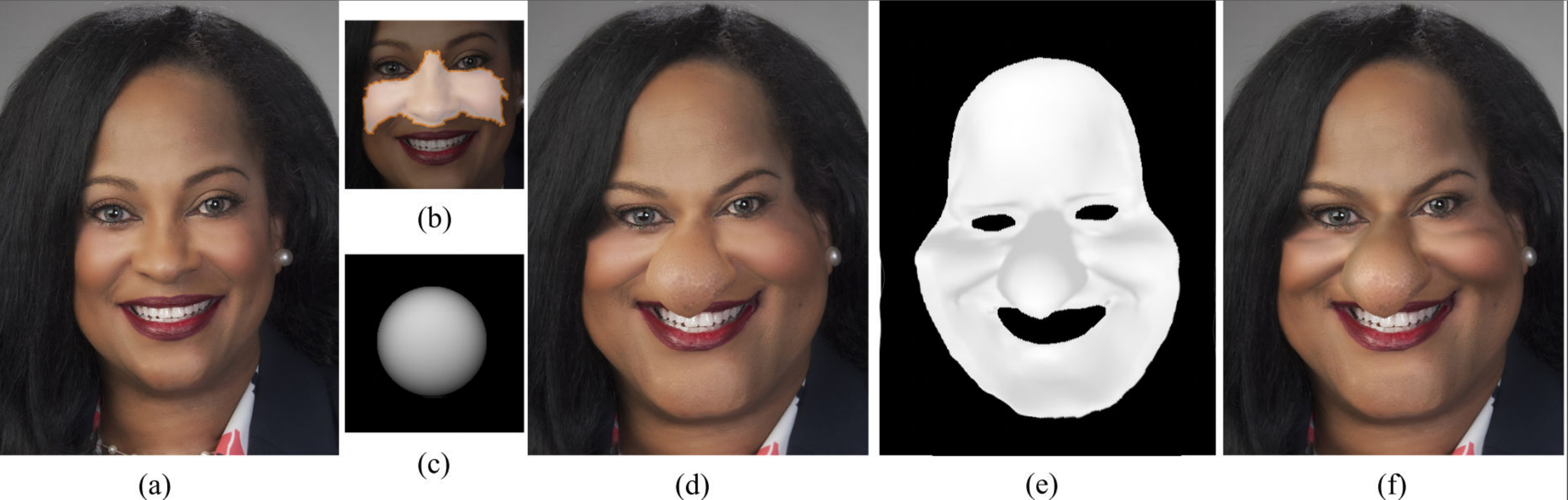}
  \caption{Input an image (a), we use the specified region as shown in (b) for global lighting estimation. (c) shows our estimated global illumination ball. (d) gives the result after our fusing step. (e) shows the optimized $\alpha$ map which is then applied on (d) and produces final result (e).}
  \label{fig:reshading_pipeline}
\end{figure}

\subsection{Fusing with Background}
To create the final $I_c$, our approach fuses $I^{f}_c$ into $I$ in three stages. This is demonstrated in Fig \ref{fig:synthesis_pipeline}.

\textbf{Warping.}
Borrowing the idea from \cite{zhou2010parametric}, we firstly do 3D-aware warping to deform $I$ in accordance with the deformation from $M$ to $M_c$. Specifically, we do regular triangulation (as shown in Fig \ref{fig:synthesis_pipeline} (a)) for the image and then perform the warping in an as-rigid-as-possible way, as the common-used strategy for image retargeting \cite{shamir2009visual}. The displacements of all front-facing vertices on $M$ are projected to force a deformation field which guides the warping. The warped $I$ is denoted as $I^{b}_c$.

\textbf{Compositing.}
To fuse $I^{f}_c$ and $I^{b}_c$, our approach follows the mechanism of \cite{dale2011video} to find an optimal cut to tailor the two images together. This is solved by using a graph cut method. After that, a poisson blending is also adopted for seamless compositing. Note that, $I^{f}_c$ does not include the parts of eyes and mouth. To generate a complete output, the content of these regions are copied from $I^{b}_c$ and warped to match the boundaries. Our final output is denoted as $I^{c}_c$ ($c$ stands for compositing).

\begin{figure}
  \centering
  \includegraphics[width=0.45\textwidth]{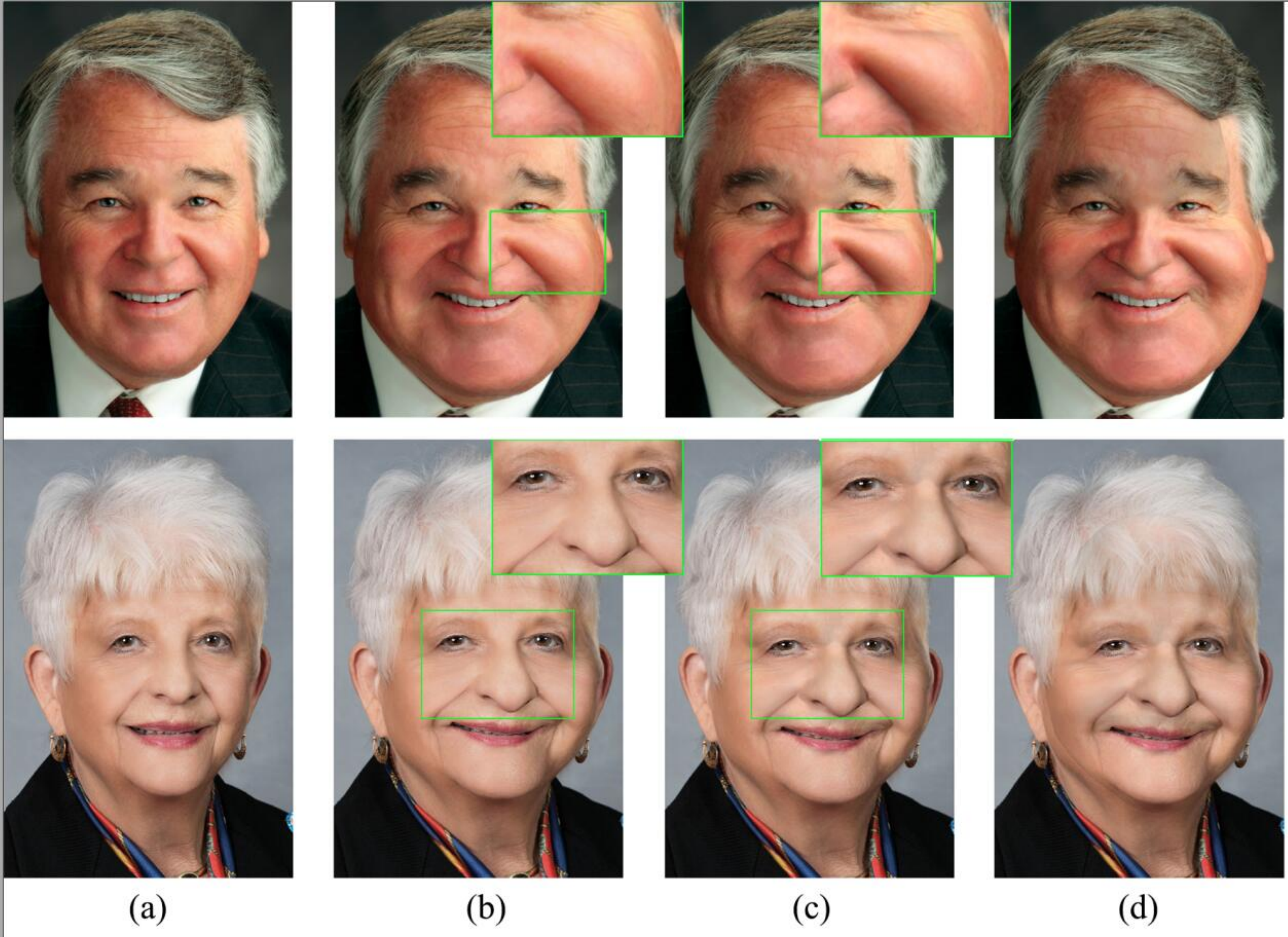}
  \caption{Input images (a), (b) and (c) show the results before and after reshading. (d) shows the results without boundary control for the $\alpha$ map.}
  \label{fig:reshading_comparison}
\end{figure}

\textbf{Reshading.}
Assuming the surface of a face is lambertian, each pixel on it can be approximated as $a*s$ where $a$ stands for albedo intensity with RGB channels and $s$ is a value representing shading. The shading is the result of lighting and geometry. This means that geometric changes brought about by exaggeration only affect the shading. To tune the shading in accordance with the geometric deformations, we only need to calculate a scaling factor $\alpha$ for each pixel. Therefore, our approach starts from global lighting estimation. We approximate the global lighting model using spherical harmonics \cite{kazhdan2003rotation}. The estimation is formulated as an optimization problem. Our formula is a simplified version of SIRFS \cite{barron2015shape} which tried to optimize the lighting, geometry and reflectance concurrently. We take the geometry information from the recovered shape as known and only treat lighting and reflectance as variables. For the energy terms, we only make use of the reflectance smoothness constraint and the illumination prior constraint. Our optimization is performed on gray-scale images. It is worth noting that, only the regions of nose and cheek are considered for lighting estimation. We argue that this not only greatly improves the efficiency but also is enough for a correct global lighting estimation. This is due to the simple albedo distribution yet rich geometry variations of these regions.

After the global lighting (denoted as $L$) is obtained, the $\alpha$ value for each pixel can be simply calculated by $L(n_c)/L(n)$ where $n$ and $n_c$ are the normal of that pixel before and after exaggeration. As a large portion of regions in $I^{b}_{c}$ has no geometric information, directly applying $\alpha$ map on $I^{c}_c$ incurs seams at the boundary. To address this issue, we resolve the $\alpha$ map by setting the $\alpha$ values of the pixels at the boundary to be $1$ and solving for the other pixels with a poisson equation. We call this procedure "boundary control". To improve the efficiency, the optimization is carried out for a downsampled version of the input image and the obtained $\alpha$ map is then rescaled to its original version before reshading. Our reshading is finally performed by multiplying $\alpha$ map to $I^{c}_c$. After that, we can create the final result $I_c$. The whole pipeline of our reshading method is shown in Fig \ref{fig:reshading_pipeline}. In Fig \ref{fig:reshading_comparison}, we use two examples to show the differences before and after reshading and the differences with and without boundary control.

\subsection{Interactively Refinement}
Our system also provides functions for further interactive refinement.

\textbf{Mouth Region Filling.}
Considering an input image whose mouth is closed but you wish to open it for editing its expressions, this will fail because of the missing content inside the mouth. To allow such manipulation, our system provides a set of mouth region templates. The users can select one of them for the content filling which is also implemented by a mesh-based warping strategy.

\textbf{Sketch-based Ear Editing.}
As the recovered shape $M$ shows severe mismatching with $I$ for the ears, our approach described above is not able to support editing ears. We provide such editing as an additional module of our refinement mode. The users can manipulate the ears by firstly interactively draw a sketch curve along the boundary of an ear and then re-draw it to provide the shape that they wish the ear to be. The ear part is then accordingly deformed using the method of \cite{eitz2007sketch}.

\section{Experimental Results}
\label{sec:results}
Our sketching system is implemented using Qt5.8. Both our \emph{caricNet} and \emph{detailNet} use tensorflow as the basic training framework and are trained with one GeForce GTX 1080Ti with CUDA 9.0 and cudnn 7.0 The \emph{caricNet} took 200K iterations with a batch size of 8 for training. This procedure spent around two days. For \emph{detailNet}, the batch size is set as 8 and the procedure also took 200K iterations. Its training procedure took about one and a half days. To evaluate each of these two networks, 10\% paired data from the dataset are randomly chosen as the validating set and the remaining ones are used for training.

\begin{figure*}
  \centering
  \includegraphics[width=\textwidth]{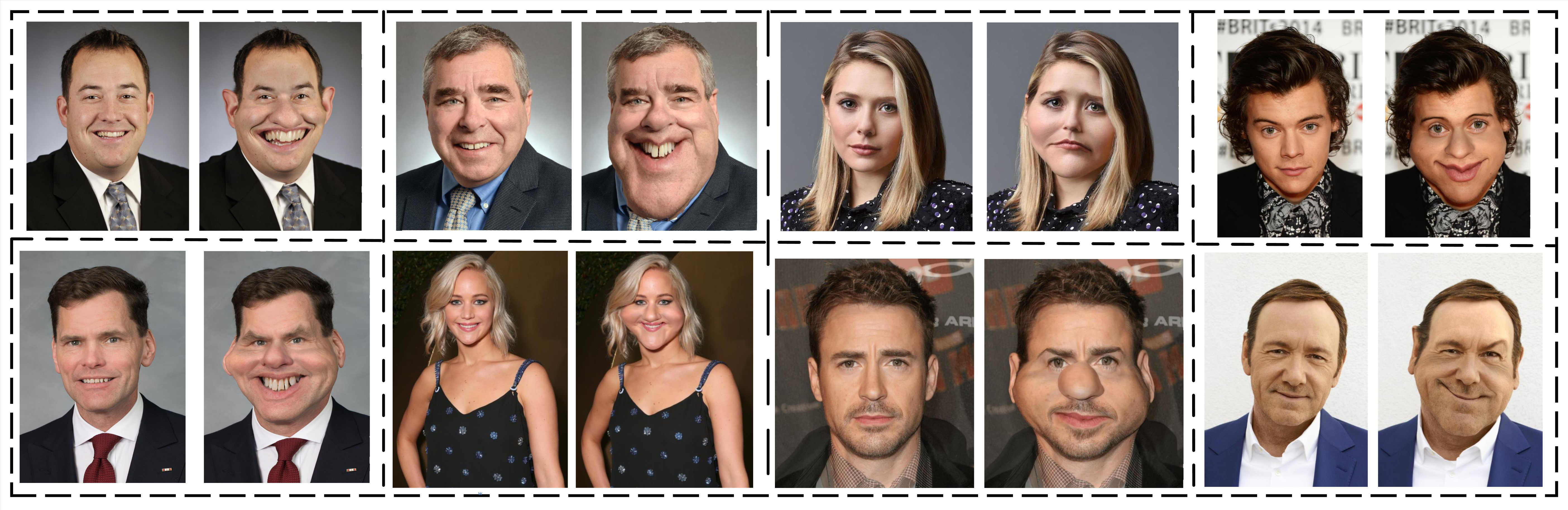}
  \caption{A gallery of results created using our sketching system. }
  \label{fig:example}
\end{figure*}

\subsection{Performance}
\begin{figure}
  \centering
  \includegraphics[width=0.45\textwidth]{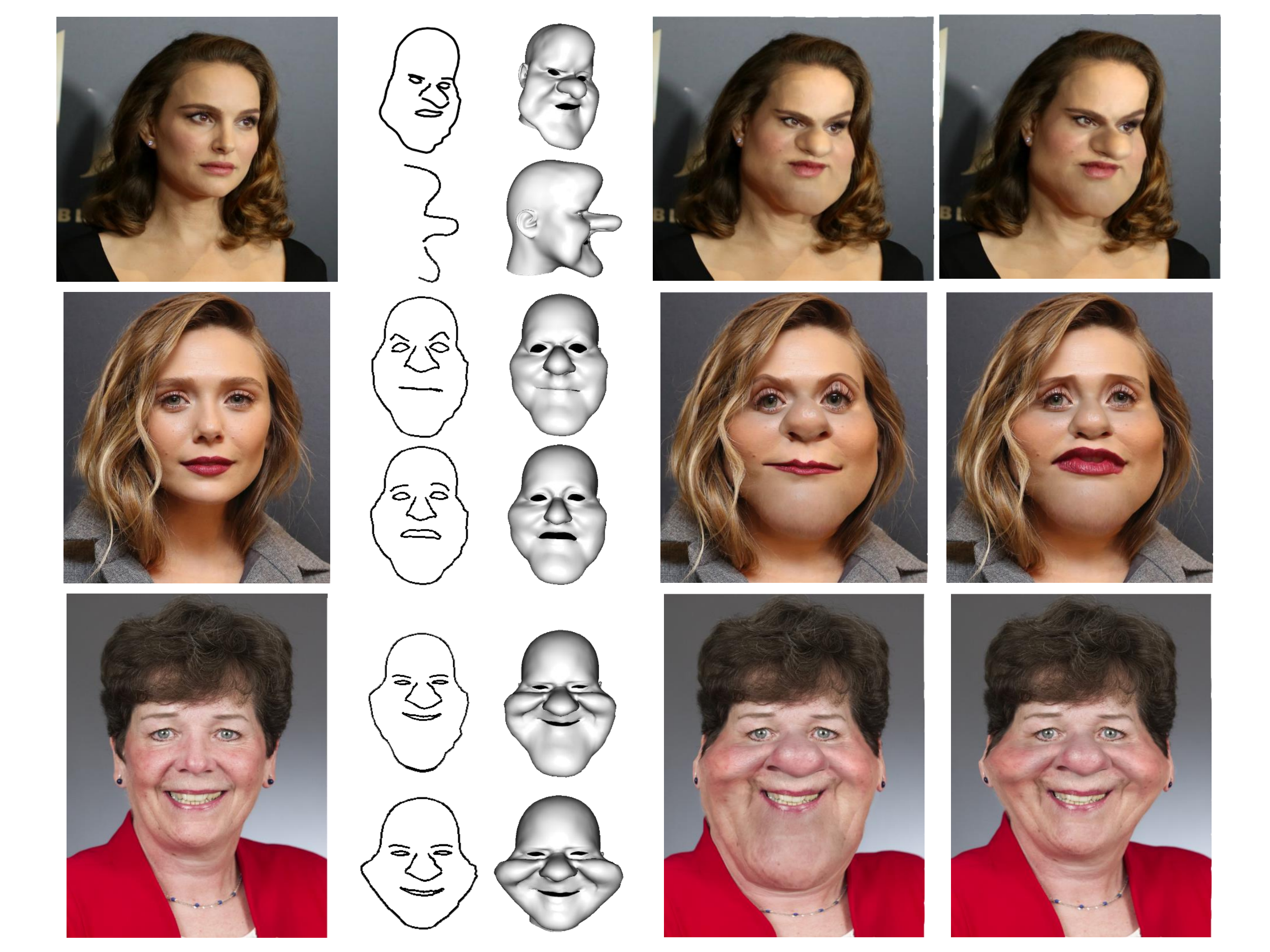}
  \caption{More results are shown. The first column gives the input photos. For each individual, the user performs two different styles of sketching. The edited sketches and the corresponding 3D exaggerated models are shown. The last two columns show caricature photos associated with the two styles. }
  \label{fig:gallery}
\end{figure}

\textbf{Qualitative Results.}
We have evaluated the performance of our caricaturing system on a large set of images (33 photos collected from internet which ranges from 720p to 1080p). The human faces in these photos are of various shapes, poses, expressions and appearances. Some of the results are shown in Fig \ref{fig:example} and Fig \ref{fig:gallery}. Fig \ref{fig:example} shows 8 examples where the input and output are placed side-by-side. In Fig , each input photo undergoes two different caricaturing styles, where the sketching and exaggerated meshes are also shown. The remaining results are listed in supplemental materials. From the qualitative results, we have the following three findings: 1) Most of the images are of detailed textures such as beard of freckles. Although our exaggeration causes severely stretching, our method well-preserves the details making the final results look photorealistic. 2) The shading effects of our results are consistent with the geometric deformations which greatly increases the stereoscopic feelings. 3) For the first row of Fig \ref{fig:gallery}, the user manipulated the depth of nose by using the side-view sketching mode. This editing is reflected by the changing of shading effects. All of these validate the effectiveness of the design of our pipeline.

\begin{figure}
  \centering
  \includegraphics[width=0.45\textwidth]{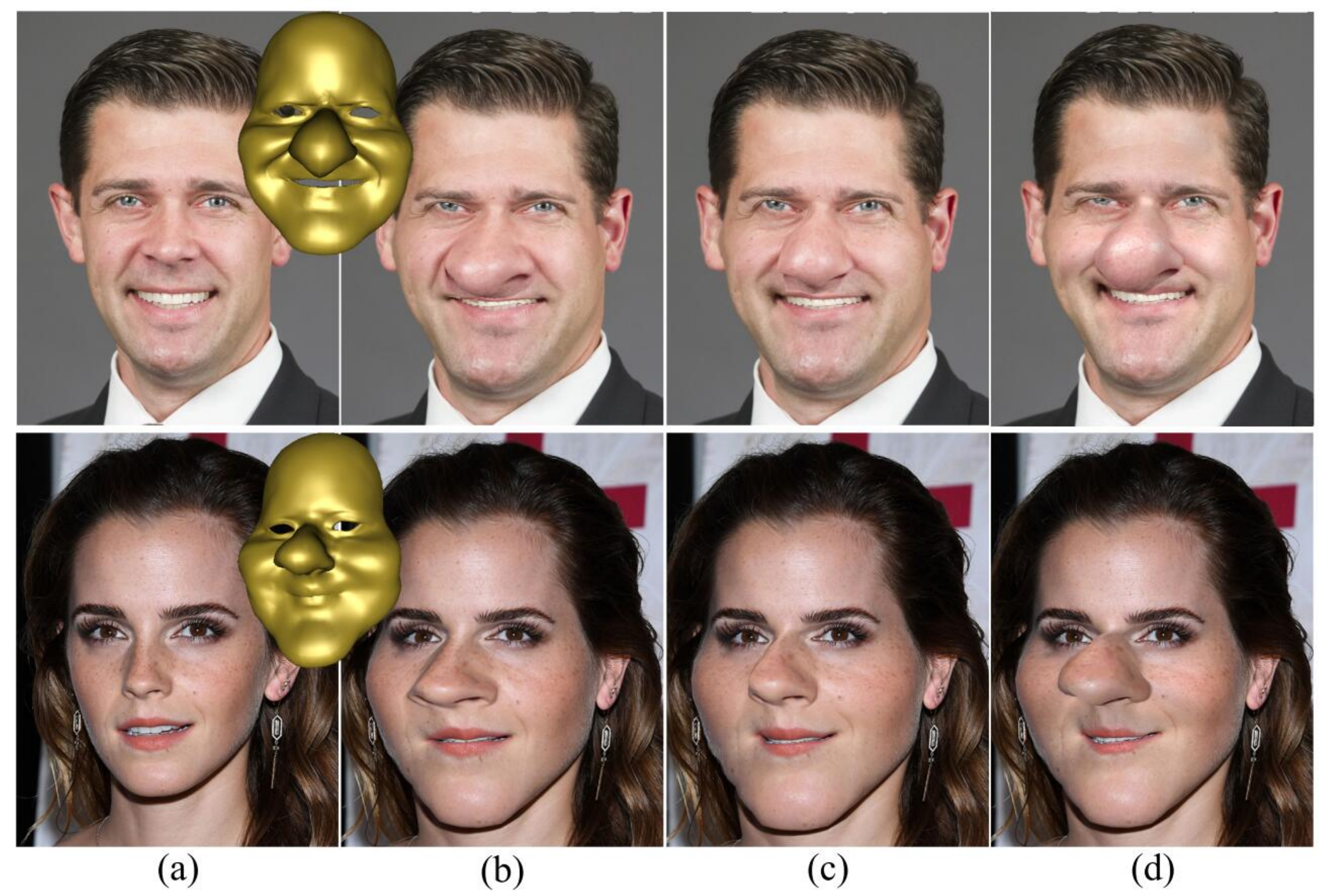}
  \caption{Comparisons on photo synthesis. (a) shows the input image. (b)-(d) show the results of \emph{2D-Warping}, \emph{3D-Warping} and ours respectively. The corresponding 3D exaggerated models are illustrated in gold. }
  \label{fig:deformcomp}
\end{figure}

\textbf{Timings.}
In summary, our framework is consisting of the following steps: 1) 3D shape recovery (using the method of \cite{cao20133d})(abbr. as \emph{shapeRec}); 2) $\lambda$ map inference from sketches (abbr. as \emph{caricInfer}); 3) caricature model reconstruction from $\lambda$ map and the position constraints of 2D sketches (abbr. as \emph{caricRecon}); 4) patch-based facial details enhancement (abbr. as \emph{detailEnhance}); 5) 3D-aware background warping (abbr. as \emph{3dWarping}); 6) image fusion using graph cut and poisson blending (abbr. as \emph{bgFusing}); 7) reshading. Both step 3) and step 5) rely on solving a sparse linear system which is implemented using CUDA. Note that, the coefficient matrices of the linear systems are pre-decomposed as in \cite{han2017deepsketch2face} to further reduce the computational cost. Both step 2) and step 4) are also carried out using a GPU. Moreover, for step 4), all of the split patches are enhanced in parallel. Although our system allows high-resolution input, both 1) and 7) can be conducted on its downsampled version. The poisson blending procedure in step 6) is accelerated by the method in \cite{farbman2009coordinates}. Note that, in our current implementation, the graph-cut is solved in CPU while we believe that this can be further accelerated using GPU. We left this as one of our further works. The average timings of each step are reported in Table \ref{tab:timing} which are calculated on the 33 images. As \emph{caricInfer} together with \emph{caricRecon} averagely cost 145ms, the sketch editing can be performed in real-time. After the users finish the editing and click a "create" button, they usually need to wait several seconds for the final results. The average waiting time is also reported as "waitTime" in Table \ref{tab:timing}.

\begin{table}[b]
\centering
\caption{Average Timings for each component of our pipeline.}
\label{tab:timing}
\begin{tabular}{|c|c|c|c|c|}
\hline
              & \textit{shapeRec}                  & \textit{caricInfer} & \textit{caricRecon}                     & \textit{detailEnhance} \\ \hline
\textit{AveT} &  67ms                                   &  102ms             &  43ms                                   &   221ms            \\ \hline \hline
              & \multicolumn{1}{c|}{\textit{3dWarping}} & \textit{bgFusing}  & \multicolumn{1}{c|}{\textit{reShading}} & \textit{waitTime}   \\ \hline
\textit{AveT} &  22ms                                   & 1,334ms            &  845ms                                  &   2,422ms            \\ \hline
\end{tabular}
\end{table}

\subsection{Comparisons}

\begin{figure}
  \centering
  \includegraphics[width=0.45\textwidth]{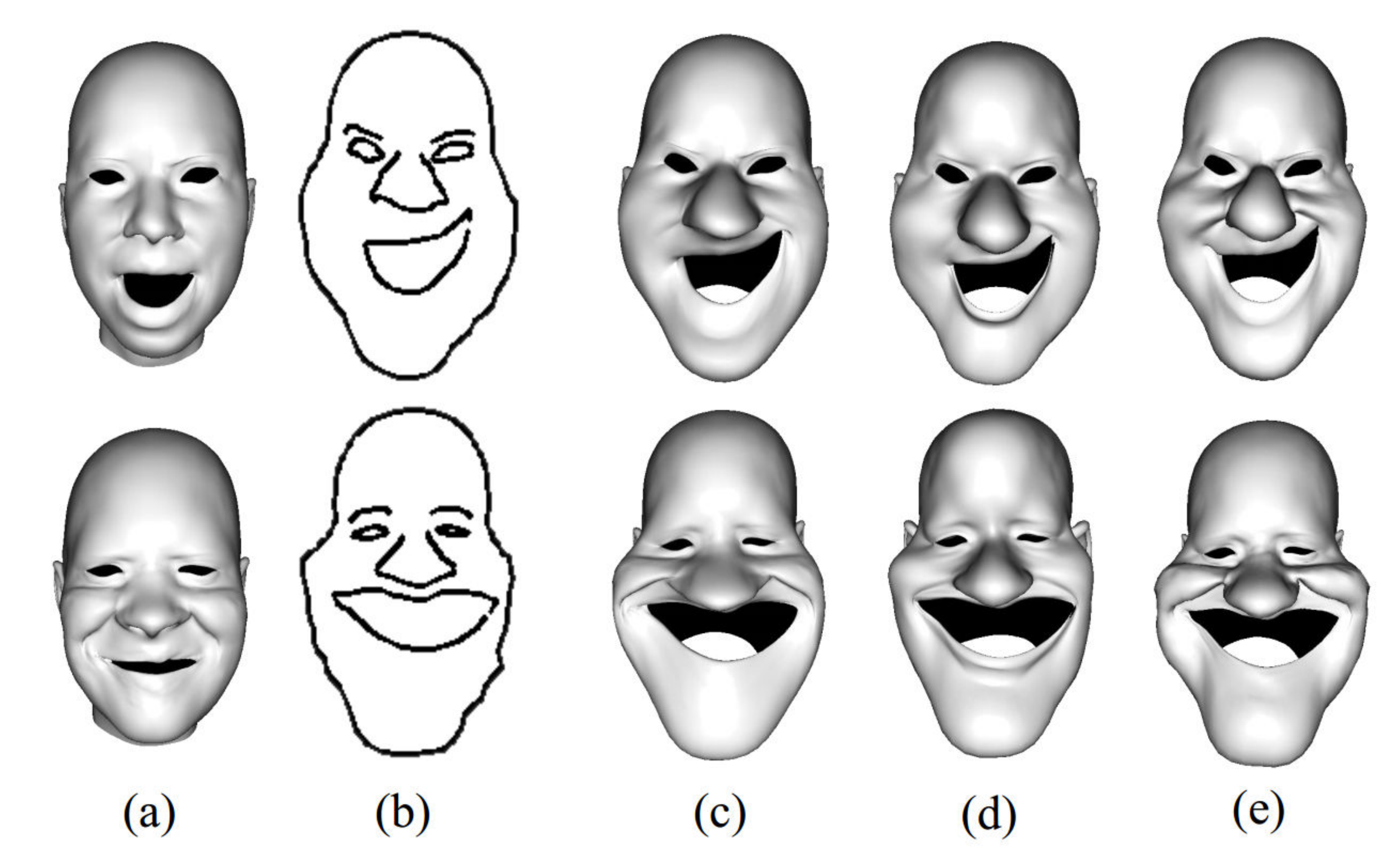}
  \caption{Comparisons on sketch-based 3D caricaturing. Input a model (a) and an edited sketch (b). (c)-(e) show the results of \emph{naiveDeform}, \emph{deepSketch2Face} and ours respectively.}
  \label{fig:3dcompare}
\end{figure}

\textbf{Comparisons on 3D Exaggeration.}
Given a 3D face model recovered from an image, there exists other ways to make caricature models from edited sketches: 1) directly do sketch-based laplacian deformation using the method in \cite{nealen2007sketch} (denoted as \emph{naiveDeform}); 2) perform deepSketch2Face \cite{han2017deepsketch2face}. Two examples are used for qualitative comparisons between our method and these two approaches which are shown in Fig \ref{fig:3dcompare}. It is obvious to find that our approach produces richer details. This is because: 1) \emph{naiveDeform} does not change the laplacian details during the deformation procedure where the sketch only provides position information for sparse vertices; 2) deepSketch2Face also makes use of a deep learning method to infer a 3D caricaturing from sketches. However, it uses a 66-dimension vector to represent caricatures while ours infers a vertex-wise field which captures larger shape space.

\textbf{Comparisons on Photo Synthesis.}
Based on the 3D models before and after exaggeration, there are also two existing methods to generate the caricature photo according to the changing of 3D shapes: 1) directly do sketch-based image deformation using the method of \cite{eitz2007sketch}(denoted as \emph{2D-Warping}); 2) perform 3D-aware image warping and output $I^{b}_c$ which is the most popular strategy as in \cite{fried2016perspective} (denoted as \emph{3D-Warping}). We also use two examples to show the qualitative comparisons between our method and these two approaches as in Fig \ref{fig:deformcomp}, where the exaggerated 3D models are also shown. As can be seen, our approach outperforms the others in two aspects: Firstly, both \emph{2D-Warping} and \emph{3D-Warping} either causes mismatching with the sketches (the first row in Fig \ref{fig:deformcomp}) or incurs distortions (the second row in Fig \ref{fig:deformcomp}). It is very challenging for these warping strategies to reach a balance because of self-occlusions happened on nose part. Secondly, the other two methods produce flat shading while ours gives rise to better lighting effects and stronger stereoscopic feelings.

\subsection{Ablation Studies}
In this subsection, we will introduce the ablation studies for our caricature inference module and details enhancement module.

\textbf{Internal Comparisons On Exaggeration Inference.}
There are several different choices for our \emph{caricNet}: 1) Instead of flattening laplacian information (i.e., $L_d$ and $L_m$) of $M^{n}$ to be input images, we can directly encode the position information of vertices as a color map for input. We denote this method as \emph{caricNet-Vertex}; 2) Before going through the Pix2Pix-Net, our method used a set of convolutional layers to follow each input map for feature transformation. Instead of using this strategy, a simple way is to concatenate all input maps together and feed them into the Pix2Pix-Net directly. This is denoted as \emph{caricNet-w/oTransform}; 3) Another variant is taking $S_c$ without flattening as input directly. To adapt our method into this setting, we firstly use Pix2Pix-Net to connect $(L_d, L_m)$ and $\lambda$ map. Meanwhile, we design an encoder to turn $S_c$ into a feature map which is then concatenated into the middle of the Pix2Pix-Net. We denote this method \emph{caricNet-w/oSketchFlatten}. We evaluate these methods and ours by using mean square error (MSE) between the output $\lambda$ map and its groundtruth. The results are reported in Table \ref{tab:caricNet} which validates our final choice is the best one. Note that, our\emph{caricNet} is only trained for frontal view sketches. We denote the network with side-view sketches embedded as \emph{caricNet} whose MSE is also reported in Table \ref{tab:caricNet}.

\begin{table}[t]
\centering
\caption{Ablations study on Exaggeration Inference.}
\label{tab:caricNet}
\begin{tabular}{|c|c|}
\hline
       & \textit{Mean Square Error (MSE)} \\ \hline \hline
 \textit{caricNet-Vertex}  & 274.0 \\ \hline
 \textit{caricNet-w/oTransform}  &  268.7 \\ \hline
 \textit{caricNet-w/oSketchFlatten} & 426.5 \\ \hline
 \textit{caricNet} & \textbf{245.1} \\ \hline \hline
 \textit{caricNet-side} & 60.1\\ \hline
\end{tabular}
\end{table}

\textbf{Internal Comparisons On Details Enhancement.}
Our facial details enhancing approach also has several variants. We firstly try to train a Pix2Pix-Net taking the whole high-resolution images as input and output their corresponding sharp photos. However, this fails using one Titan X GPU with 12 GB memory. This validates the necessity of using patch-based approach. We further evaluate the methods with or without using residual. The average MSE without residual is 27.1 while ours method produces 21.0. This also validates the superiority of our design.

\section{Conclusions and Discussions}
\label{sec:conclusion}
In this work, we have presented the first sketching system for interactively photorealistic caricature creation. Input a portrait photo with a human face in it, with our system, the users can do the caricaturing by manipulating the facial feature lines based on their personal wishes. Our system firstly recovered the 3D face model from the input image and then generated its caricatured model based on the edited sketches. Our 3D exaggeration is conducted by assigning the laplacian of each vertex a scaling factor. For sake of building the mapping between the 2D caricature sketches and vertex-wise scaling factors, a deep learning architecture is exploited. To do this, we proposed to flatten the information on meshes into a parametric domain and encode both the 3D shape and 2D sketch to a set of images. A variant of Pix2Pix-Net \cite{isola2017image} is thus utilized for translating such 2D maps to the vertex-wise scale map. Based on the created caricatured model, our photo synthesis followed several steps. Firstly, we did facial detail enhancement which aimed to infer the missing details for the blurry regions caused by stretching of meshes. A deep learning architecture is also adopted for this inference. After that, we fused the projected textured image with the warped background together and applied a reshading operation to obtain the final result. The qualitative comparisons show that our framework outperforms all existing methods and the quantitative results of the ablation studies also validate the effectiveness of our network design.

\begin{figure}
  \centering
  \includegraphics[width=0.45\textwidth]{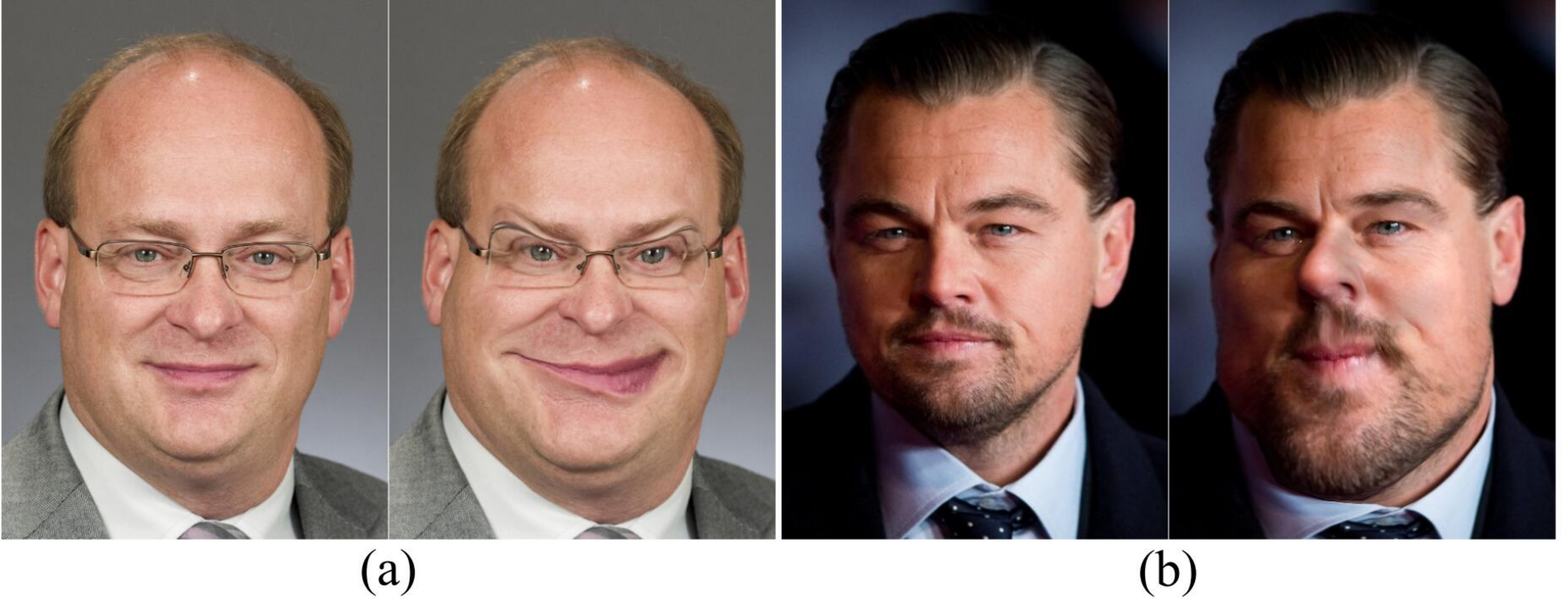}
  \caption{Limitations of our method.}
  \label{fig:limitation}
\end{figure}

\textbf{Limitations.}
Our system still has limitations for two challenge scenarios. At first, for the facial images with accessories such as glasses as shown in Fig \ref{fig:limitation} (a), our approach causes distortions. This is due to the lacking of 3D information of the glasses. Thirdly, our reshading method only captures the global lighting which makes it difficult to deal with complicated lighting environments. Taking note that the example shown in Fig \ref{fig:limitation} (b), our approach produces wrong shading effects as a result of the wrong estimated global lighting model. 

\vspace{-1mm}
\bibliographystyle{IEEEtran}
\bibliography{caricatureshop_tvcg}

%


\end{document}